# Is ChatGPT a game changer for geocoding - a benchmark for geocoding address parsing techniques*

Zhengcong Yin
Texas A&M University
College Station, TX
yinzhengcong@tamu.edu

Diya Li
Texas A&M University
College Station, TX
diya.li@tamu.edu

Daniel W. Goldberg
Texas A&M University
College Station, TX
daniel.goldberg@tamu.edu

## ABSTRACT

The remarkable success of GPT models across various tasks, including toponymy recognition motivates us to assess the performance of the GPT-3 model in the geocoding address parsing task. To ensure that the evaluation more accurately mirrors performance in real-world scenarios with diverse user input qualities and resolve the pressing need for a 'gold standard' evaluation dataset for geocoding systems, we introduce a benchmark dataset of low-quality address descriptions synthesized based on human input patterns mining from actual input logs of a geocoding system in production. This dataset has 21 different input errors and variations; contains over 239,000 address records that are uniquely selected from streets across all U.S. 50 states and D.C.; and consists of three subsets to be used as training, validation, and testing sets. Building on this, we train and gauge the performance of the GPT-3 model in extracting address components, contrasting its performance with transformer-based and LSTM-based models. The evaluation results indicate that Bidirectional LSTM-CRF model has achieved the best performance over these transformer-based models and GPT-3 model. Transformer-based models demonstrate very comparable results compared to the Bidirectional LSTM-CRF model. The GPT-3 model, though trailing in performance, showcases potential in the address parsing task with few-shot examples, exhibiting room for improvement with additional fine-tuning. We open source the code and data of this presented benchmark[1] so that researchers can utilize it for future model development or extend it to evaluate similar tasks, such as document geocoding.

## 1 INTRODUCTION

Geocoding is the process of converting address descriptions into geographic coordinates [12] and has been widely used as a data-processing step in various domains to conduct spatial analysis, from enabling efficient urban planning to advancing public health [21, 28, 29, 40, 42]. However, the validity of conclusions of studies that employed geocoding as part of their workflow can be largely impacted by the quality of geocoded data. [18, 31, 39]. Although every step in a geocoding process can accumulate errors in the final output [8, 12], **address parsing**, which is to extract address components (shown in the Figure 1) from the address description input by users, plays a profound role in determining the quality of geocoded data. This is because outputs from the address parsing process are used to assemble query strings to retrieve matching candidates for further calculation and ranking to derive final geocoded outputs [8, 9, 38]. For example, if an address parser mistakenly recognizes the '116 S' as the house number, the geocoding engine can not extract the correct geocode output from the reference dataset by employing '116 S' as a house number search criterion. Moreover, geocoding input is proven to be error-prone, containing syntactic or semantic errors [3, 17]; such quality of user input demands an address parser to handle them appropriately to ensure the quality of geocoded output.

Recently, significant breakthroughs have been witnessed in Large Language Models (LLMs) and their applications in Natural Language Processing (NLP). Models like GPT-3 [1] are making waves by setting new benchmarks in various tasks [35] and by shifting the model training workflow [7]. Researchers in the geospatial domain have also evaluated the capability of Generative Pre-trained Transformer (GPT) models in handling toponymy recognition and location description recognition tasks and have shown promising results [26]. This raises the question that ***can GPT-based models make a difference in the task of address parsing?***

Yet, there has been minimal effort in evaluating the performance of GPT-based models for address parsing or geocoding, and the favorable evaluation outcomes presented in [26] can not provide too many insights into the performance of GPT-based models on geocoding address parsing, as their evaluation framework [15] is not designed for the address geocoding task, and input for toponymy resolution tasks, which contain place names in short text messages, is different from the input for geocoding: a postal address description. In fact, in the realm of geocoding, a "gold-standard" benchmark dataset that can fully evaluate geocoding systems is highly demanded [18]. Compared to the magnitude of human input errors, input datasets in existing geocoding evaluation frameworks only contain relatively simple misspellings [3, 17]. We argue that evaluation data used in existing work did not fully reflect the quality of geocoding input in real scenarios, making the performance of a geocoding system and address parser remain uncertain when facing (erroneous) input in reality.

Therefore, in this work, we present a benchmark that is specific to evaluate geocoding address parsing techniques using synthesized low-quality input by mining human input patterns from real geocoding system logs, and we evaluate address parsing performance using GPT-3 model and compare to other transformer-based and recurrent neural network-based address parsing methods.

---

[1]https://github.com/zhengcongyin/Geocoding-Address-Parsing-Benchmark

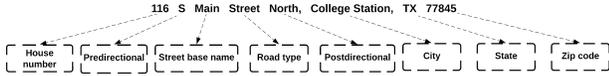

**Figure 1: USPS standard address components of a postal address description**

The contributions of this work can be summarized as follows.

- A benchmark dataset that contains diverse address descriptions (e.g., highway and grid style) covering all U.S. states and 21 input errors and variations is generated by mining real geocoding system logs. A data processing pipeline is developed to analyze input errors and variations occurring in different address components from real user input, and then the harvested inject errors and variations using these identified patterns to synthesize low-quality geocoding input. To the best of our knowledge, this is the first publicly released annotated low-quality geocoding input dataset for U.S. addresses with such magnitude of coverage and error/variation.
- Address parsers built upon five different models (i.e., GPT-3 model, transformer-based model, and LSTM-based model) are evaluated by synthesized low-quality address input with different errors to reflect their performance when facing various input qualities in real scenarios. These evaluation results can provide insights into the potential capabilities of each model, especially the GPT-3 model, for further fine-tuning or enhancement.
- The proposed benchmark, encompassing benchmark datasets and address parsing methods, is available as open source and can be accessed at Github[2]. Researchers could use the benchmark dataset for other geospatial text processing tasks or use the evaluation results as baselines for future development and experimental comparisons. This proposed framework can be extended to synthesize language- or county-specific low-quality input to evaluate address parsing or geocoding systems in different countries.

The remainder of this paper is organized as follows. Section 2 summarizes recent work on address parsing techniques in geocoding and Name Entity Recognition in other domains. Section 3 describes the design details of the proposed benchmark, including the approach to synthesize the low-quality geocoding input, evaluated address parsing techniques, and evaluation metrics. In Section 4, we illustrate the evaluation outcomes and discuss the results. We conclude this paper with potential avenues for future work in Section 5.

## 2 RELATED WORK

Geocoding address parsing is a domain-specific Named Entity Recognition (NER) task that has received extensive research attention. In previous research endeavors, the primary competitor has centered on algorithms designed to improve address parsing capabilities. In the initial stage, these parsing algorithms were predominantly built on rule-based and statistical methodologies. Rule-based approaches usually leverage the format of local address schema and its hierarchy to determine the sequence of labels for a given address input [8]. Typically, a tire or tree-based data structure is used to mimic the hierarchy of address systems, string matching (i.e., forward/backward string method), beam search, heuristic search strategies, and fine-state machines are used to explore the possible label sequences for addresses. Given that rule-based methods heavily rely on address system rules and lexicons to recognize certain address components (e.g., road types), the variation of user input in terms of quality and descriptions could easily result in the "Out Of Vocabulary" issue. Later, statistical-based address parsing represents a learning and tagging process, as an annotated corpus is required for training, and sequence tagging algorithms make decisions for each label. Two popular models, Hidden Markov Models (HMMs) and conditional Random Fields (CRFs) have been used to build address parser [2, 33] and achieved SOTA at that time. To augment the coverage of the state transition matrix for variations, [2] has enhanced the training data to contain intentionally manipulated addresses. Later, the hybrid-based address parser that combines the rule-based and statistical-based approaches [23] has shown better parsing performance. In recent years, research has shifted towards using neural networks and LLMs as the foundational framework for building address parsers[13, 14, 27, 30, 36], given their proven success in NER tasks across various domains [20]. Another avenue of research related to address parsing involves reducing the need for annotated data [4] or predicting noisy tokens in geocoding queries[34].

Given that address descriptions and formats differ among countries [36], the aforementioned studies are using input address descriptions from the address system specifics to their study areas, including U.S. [9], China [22], Japan [24], and India [27]. However, the lack of a standardized evaluation dataset for each individual address system complicates the direct comparison of the experimental results of different studies targeting the same country. Our work extends the existing works by presenting a unified evaluation framework, including a benchmark dataset, evaluation procedures, and evaluation metrics created specifically to assess geocoding address parsing. The benchmark dataset, which accounts for the heterogeneity in address formats and encompasses a wide range of input errors/variations, is publicly released to facilitate future research investigations.

## 3 BENCHMARK DESIGNS

This section describes how benchmark datasets are generated, the selection of evaluated models, and the metrics to access address parsing performance.

### 3.1 Benchmark Dataset

Figure 2 depicts the workflow to generate the benchmark dataset, namely, the low-quality geocoding input dataset. This workflow contains three major steps: (1) extracting ground-truth dataset; (2) building *address component error injector* that can generate common

---
[2]https://github.com/zhengcongyin/Geocoding-Address-Parsing-Benchmark

geocoding input error and variations; (3) synthesizing low-quality geocoding input.

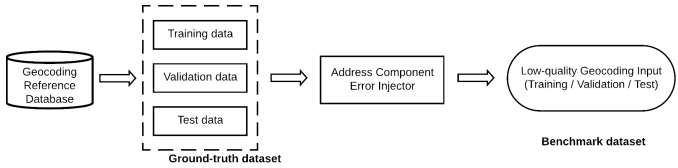

**Figure 2: Benchmark dataset processing workflow**

*3.1.1 Ground-truth data.* The ground-truth data is generated by extracting from reference datasets, as reference datasets are the single source of truth for geocoding systems to perform the retrieval processing to derive final outputs. We extracted address description from the Navteq 2016 address point reference datasets [3] used by Texas A&M geocoding platform[4], because this dataset has been utilized by other studies [37, 38], and every address description in this reference dataset has already been segmented and aligned with a USPS standard address component label shown in Figure 1. To ensure the diversity of address descriptions across the U.S., we first get the unique combination of address components except for house number (i.e., street name, predirectional, postdirectional, city name, and postal code) from each U.S. state and the District of Columbia, meaning that we obtained one address description from every street from all U.S. 50 states and D.C to formalize a unique address description dataset. Then, we further split this unique address dataset into three smaller datasets designated for training, validation, and testing procedures in this benchmark. The testing dataset is generated by extracting one address description from every pair of state and postal codes in the U.S., resulting in a dataset of 30,622 addresses. To obtain the training and validation datasets, we first exclude the testing dataset from the unique address collection; we then randomly select up to 9 addresses from every unique combination of city, state, and postal code from the unique address collection and put the first two addresses and the last address into the training and validation dataset, respectively, when applicable. In the end, the training and validation datasets have 148,173 and 60,522 addresses, respectively, and all address descriptions in the training, validation, and test datasets are mutually exclusive.

*3.1.2 Address component error injector.* To synthesize low-quality geocoding input, we build an *address component error injector* to randomly generate errors and variations based on human input patterns. To capture such patterns for geocoding input, we firstly extract three-month geocoding transactions from Texas A&M geocoding platform[5] and only keep these inputs, which cannot lead to full matching scores (i.e., the reference data can only partially match with the input.) In total, we obtained roughly 30 million input queries. Next, we iterate each input and compare it to its corresponding reference data to detect input errors and variations. Since user input and matched reference data have already been segmented based on address components to seek a match by the geocoding platform, input errors can be found by aligning user input address descriptions with their corresponding description in address reference datasets. For example, if the city name is missing in the input compared to the corresponding reference data, the error of *omission* is detected. While iterating the historical user input data, we collect sets of mismatched input samples per each address component and then further distill to get cases of commonly used abbreviations and common substations for each address component. Totally, we identify 21 errors and variations on different address components listed in Table 1.

[3] https://www.here.com/en/navteq
[4] https://geoservices.tamu.edu/
[5] https://geoservices.tamu.edu/

Lastly, we create the logic to generate these identified errors and variations by aligning and comparing between the segmented user input and the segmented reference data. Addition or omission errors can be generated by reversing the process of how these errors/variations are detected. For instance, a directional addition error can be identified by comparing the user input and the reference data; such an error can be synthesized by adding a directional to an address record in the reference data. For typographic errors, we employ the same mechanism of Freely Extensible Biomedical Record Linkage (FEBRL) [3] to randomly swap, delete, insert, or replace a character. We quantify the degree of typographic error by edit distance and set the probabilities of typographic errors with edit distance 1 or 2 to be the same. As for error/variation of abbreviation and substitution, we leverage the collected common cases to reproduce the error/variation, for example, replacing *Los Angeles* by *LA* for a city name input.

*3.1.3 Low-quality geocoding input for benchmarks.* The last step is synthesizing low-quality geocoding input used as the benchmark dataset to access address parsing techniques. We apply the address component error injector to the training, validation, and test ground-truth datasets obtained in Section 3.1.1. Specifically, we set the probability of injecting errors/variations to an address record in these three split datasets to 0.5, and the ratio of injecting two or one error/variation is 7:3. Every address component has the same chance to be manipulated to contain an error/variation. It's worth noting that we only inject errors that are applicable to an address record. For example, if an address is an ordinal number street such as "5th Avenue", it is applicable to have the error of ordinal number suffix omission to become "5 Avenue". We intentionally reduce the postal code digits mismatched error to prevent it from dominating synthesized errors. To this end, training, validation, and test datasets all contain address records with zero, one, or two errors/variations, as summarized in Table 2. The distribution of each error/variation for each dataset is summarized in Figure 3. To label these three datasets for training and evaluation, we employ the IOB (Inside–outside–beginning) tagging scheme to assign the corresponding label to each chuck segmented by white space. For example, the city name *Los Angeles* would receive two labels: *B-CITY* and *I-CITY*.

## 3.2 Baseline models

The following section provides an overview of the baseline models utilized to build address parsers in our experiments, each representing significant strides in the field of NLP.

Table 1: Geocoding input errors and variations

| Address component | Error/Variation | Example |
| --- | --- | --- |
| House number | Omission | **1600** Main St → Main St |
| Pre-/Post-directional | Omission | **East** Main St → Main St |
| | Pre/Post-Direction swap | E Main St NW → **NW** Main St **E** |
| Street base name | Typo (edit distance 1) | Main St → **Man** St |
| | Typo (edit distance 2) | Main St → **Mian** St |
| | Number suffix omission | 5th Ave → **5** Ave |
| | Spanish prefix omission | **La** Brea Ave → Brea Ave |
| | Space omission | Memory Hill → **Memoryhill** |
| | Space addition | Reachcliff → **Reach Cliff** |
| | Partial abbreviation | Warm Mountain → Warm **Mtn** |
| Road type | Omission | Main **St** → Main |
| | Valid road type substitution | Main St → Main **Ave** |
| | Invalid road type substitution | Main St → Main **St St** |
| City | Omission | **Houston**, TX 77845 → TX 77845 |
| | Typo (edit distance 1) | Austin → **Austiun** |
| | Typo (edit distance 2) | Luverne → **Luvre** |
| | Direction addition | Houston → **South** Houston |
| | Direction omission | **North** Little Rock → Little Rock |
| | First character abbreviation | Los Angeles → **LA** |
| | Space addition | Redlands → **Red Lands** |
| State | Omission | Houston, **TX** 77001 → Houston, 77001 |
| Postal code | Omission | Houston, TX **77001** → Houston, TX |
| | Any digits mismatched | 77845 → **77843** |

Table 2: Frequency of address records with different quality in the benchmark dataset

| Subset | Total | No error/variation | One error/variation | Two error/variation |
| --- | --- | --- | --- | --- |
| Training | 148,173 | 74,086 | 51,898 | 22,189 |
| Validation | 60,522 | 30,230 | 21,247 | 9,045 |
| Test | 30,622 | 15,286 | 10,736 | 4,600 |

**Bidirectional LSTM-CRF** [16]: The Bidirectional LSTM-CRF model combines the strengths of both Bidirectional Long Short-Term Memory (Bi-LSTM) and Conditional Random Fields (CRF) for sequence labeling tasks. Bi-LSTM, a type of Recurrent Neural Network (RNN), is capable of capturing the context from both directions of a sequence and hence is widely used for NLP tasks. CRF, on the other hand, is a statistical modeling method often used for structured prediction. In the context of NLP, CRFs are used to predict the most likely labels for a sequence of words. The Bidirectional LSTM-CRF model leverages the Bi-LSTM to extract complex features from input sequences, and then use the CRF to predict the optimal labeling sequence, considering both the input sequence and the correlation of labels, resulting in state-of-the-art performance on various sequence labeling tasks.

**BERT** [5]: Bidirectional Encoder Representations from Transformers (BERT), developed by Google, revolutionized the NLP landscape by introducing a novel pre-training objective known as Masked Language Model (MLM). This objective allows BERT to understand the context of a word by considering both its preceding and following words, a significant departure from previous models that only captured unidirectional contexts. Pre-trained on a substantial corpus of unlabelled text, including the entirety of Wikipedia and the Book Corpus [6, 41], BERT has shown remarkable performance across a variety of NLP tasks. We choose to use the standard *bert-base-uncased* model which contains 110M parameters.

**roBERTa** [25]: roBERTa, a variant of BERT introduced by Facebook, further refines the pre-training process. It eliminates the next-sentence pretraining objective, modifies several key hyperparameters, and leverages larger mini-batches and learning rates. Additionally, roBERTa is trained on an augmented version of the BookCorpus dataset [6, 41], leading to improved performance over BERT in several benchmark tasks. We choose to use the standard *roberta-base* model which contains 125M parameters.

**DistilBERT** [32]: As a distilled variant of BERT, DistilBERT represents an effort to optimize the balance between model performance and resource efficiency. DistilBERT is 60% smaller in size, six times faster, yet retains 95% of BERT's performance. This is achieved through a process known as distillation, where a smaller model (the student) is trained to mimic the behavior of a larger model (the teacher). We choose to use the standard *distilbert-base-uncased* model which contains 67M parameters.

**GPT-3** [1]: GPT-3, also known as ChatGPT, the successor to GPT-2 and also developed by OpenAI, is an autoregressive language model with a staggering 175 billion machine learning parameters.

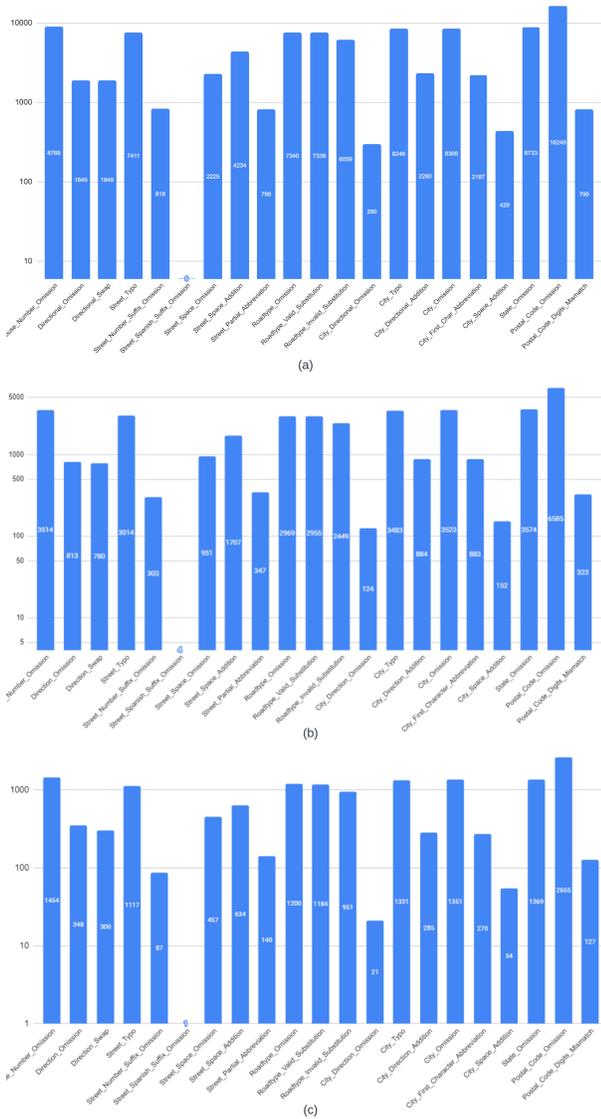

Figure 3: The distribution of synthesized geocoding input errors/variations in training (a), validation (b), and test (c) datasets

GPT-3's size and complexity enable it to excel in tasks involving the generation of long, coherent text passages. In addition to this, GPT-3 exhibits remarkable proficiency in translating between languages, answering questions, summarizing text, and more, making it one of the most versatile language models to date.

### 3.3 Evaluation metrics

Given the task of geocoding address parsing is to segment the input address description and assign a corresponding address component label to each segmentation based on the USPS address standard, we quantify the parsing performance by the standard NER evaluation metrics, namely, the precision and recall, and the F1 score (i.e., the harmonic mean of precision and recall) of every annotated label. Such a measurement indicates a parsing model's capability to recognize all address components correctly. Since the output from geocoding address parsing is to build a query string to retrieve and rank matched candidates, it's possible that not all address components would be used to build queries, and some address components are more important than others, depending on how the matching component of a geocoding system is built. To this end, we further calculate a score (denoted as *parsing score*) based on the weight of each address component used by Texas A&M geocoding platform[6] using Equation 1 as follows.

$$\text{Parsing Score} = \sum W_{address} \times F1_{address} \quad (1)$$

where, $W_{address}$ and $F1_{address}$ represents the weight and F1 score of every address component, respectively. The weight of each address component (shown in Table 3) is obtained from Texas A&M geocoding platform[7] given its performance in [10, 11].

Table 3: Weight of each address component

| Address Component | Weight |
|---|---|
| House number | 20 |
| Predirectional | 7 |
| Street base name | 45 |
| Road type | 10 |
| Postdirectional | 4 |
| City | 17 |
| State | 1 |
| Zip code | 45 |

## 4 EXPERIMENT RESULTS AND DISCUSSION
### 4.1 Model Implementation

Since the GPT-3 model generates output via user prompt, we conducted the NER task via the promptify library [8]. This library sends a structured input to LLMs, which is equivalent to asking a properly structured question that would help these GPT-3 understand the question better. The API version we used is *gpt-3.5-turbo* [9]. We supplied three examples to the GPT-3 model to help it understand the expectations for the output, as we found the output under the zero-shot scenario is suboptimal. These three examples listed below are randomly selected from the training dataset, containing pre-directional, post-directional, and no directional.

(1) *467 W BROOKWOOD CIR OZARK AL 36360*
(2) *27195 DORY RD W SALVO NC 27972*
(3) *118 LUKE HICKS RD HAZEL GREEN AL 35750*

The three transformer-based models, along with the Bidirectional LSTM-CRF model, were implemented utilizing the Pytorch framework. These transformer-based models were built using the hugging face library [10]. We trained these models using an Adam optimizer

---
[6]https://geoservices.tamu.edu/
[7]https://geoservices.tamu.edu/
[8]https://github.com/promptslab/Promptify
[9]https://platform.openai.com/docs/models/gpt-3-5
[10]https://huggingface.co/docs/transformers/index

[19], a popular choice for training deep learning models due to its efficiency and low memory requirements. The initial learning rate was set to 0.00002, with the linear learning rate schedule type. The Adam optimizer was configured with beta1 and beta2 parameters set to 0.9 and 0.999, respectively. The dropout is set to 0.5, as we observed that the default dropout can easily lead to over-fitting in the initial stage of this experiment. The batch size is set to 30. The Bidirectional LSTM-CRF model was implemented on an open-sourced work[11]. Specifically, we employed GloVe.6B.100d [12] for word embedding to fed into neural network, the stochastic gradient descent optimizer with a learning rate of 0.1, the hidden size of an LSTM cell of 200, and a batch size of 10. We added an IOB label constraint for transition parameters to enforce valid transitions.

### 4.2 Experiment Settings

This experiment aims to compare the different baseline models' performance on the task of geocoding address parsing. To have a fair comparison, we utilized the same datasets, run-time environment, and training/evaluation procedures to ensure any differences in performance could be attributed to the models' architecture and capabilities rather than external factors. Among these five baseline models, the four (i.e., the Bidirectional LSTM-CRF model and three transformers-based models) require a training process, whereas the GPT-3 model does not need to train, as we directly leveraged the gpt-3.5 turbo API to conduct NER inference for address parsing. Thus, we first trained and evaluated these five baselines using training and validation datasets to get their trained models; we then applied these trained models and the GPT-3 model to the test dataset to compare their performance. We set the training epoch to be 25, as the preliminary experiment indicated the evaluation loss was less than 0.001. Each training model was evaluated on the validation dataset at the end of each epoch, and their evaluation loss was recorded. This allowed us to monitor the models' learning progress and adjust the training parameters if necessary. All training and evaluation processes were conducted on Google Colaboratory with the Tesla V100 GPU.

### 4.3 Results and discussion

Figure 4 presents the trajectories of training and validation loss for the baseline models throughout the entirety of the experimental processes. The roBERTa model's validation performance is initially high but experiences a rapid decrease as training progresses. In contrast, the other models exhibit a steady validation loss throughout the entire process. Most models reach convergence around the 20-epoch mark. Notably, the DistilBERT model stands out for its faster convergence rate than the other models. Having trained these four models, we then tested them alongside the GPT-3 model using the same test dataset detailed in Section 3.1. The evaluation results of the five evaluated baseline models are presented in Table 4, illustrating their effectiveness in recognizing and extracting individual address elements and the overall performance.

Across all address components, the Bidirectional LSTM-CRF model consistently demonstrates superior or comparable performance to the other models. For instance, in identifying the house number, this model achieved the highest F1 score of 0.99977, marginally surpassing the performance of roBERTa (0.99976) and BERT (0.99963). Its superiority is also evident in parsing the state and postal code components, where it yielded an F1 score of 0.99993 and a perfect score of 1.00000, respectively. The BERT model exhibits robust performance across all tasks, with its performance closely trailing that of the Bidirectional LSTM-CRF model. It performed particularly well in identifying the house number and postal code, with F1 scores of 0.99963 and 1.00000, respectively. Notably, the roBERTa model, while generally performing well, exhibited a slight drop in performance when parsing the postdirectional component, with an F1 score of 0.94003. The Bidirectional LSTM-CRF model also has the highest Parsing Score, indicating that it not only performs well in parsing each address component but also excels in parsing the components that carry the most weight in the geocoding process. This is significantly lower than the scores achieved by the other models for this task. On the other hand, the DistilBERT model's performance was consistently high across all tasks, with its lowest F1 score being 0.96771 for the postdirectional component. Its performance was particularly strong in parsing the house number and postal code, achieving F1 scores of 0.99970 and 1.00000, respectively. The GPT-3 model, however, displayed a relatively lower performance compared to the other models. While it performed reasonably well in parsing the house number, state, and postal code with F1 scores of 0.98810, 0.97505, and 0.97851, respectively, it struggled significantly with the postdirectional component, achieving an F1 score of 0.42917, which is markedly lower than the scores of the other models.

The Bidirectional LSTM-CRF model consistently outperforms or matches the other models across all address components. This superior performance could be attributed to the inherent strengths of this model. The Bidirectional LSTM-CRF model combines the advantages of both bidirectional LSTM and conditional random fields, which allows the model to capture context from both past and future input while CRF can make the most of the sentence-level tag information, making it a powerful model for sequence labeling tasks such as NER. The BERT model and its variant, while performing robustly across all tasks, fall slightly behind the Bidirectional LSTM-CRF model in terms of performance. This could be due to the fact that while BERT is a powerful model, it is pre-trained on a masked language model and next-sentence prediction tasks, which may not be perfectly aligned with the NER tasks in address parsing. On the other hand, the pre-trained models have been trained on large amounts of data, their performance could potentially be improved with hyperparameter tuning to optimize them for the specific task of address parsing. This could involve adjusting parameters like learning rate and batch size, adding or removing layers, or changing the number of hidden units, among other things. However, its strong performance in identifying house numbers and postal codes suggests that it is still a valuable tool for these tasks. As a generative model, GPT-3 demonstrates a lower performance compared to others. One of the potential reasons for that is the GPT-3 generates output based on the context provided by the prompt. Therefore, the way the prompt is framed can significantly affect the model's performance. The other reason could be the selected few-shot learning examples used by the GPT-3 model were completely error-free. It would be interesting to compare the impact of different few-shot learning examples on the GPT-3 model performance. It's worth noting that the hyperparameters of evaluated models come from common settings used by other studies, as the main scope of this paper is to provide a solid foundation to facilitate future model evaluations. Fine-tuning hyperparameters for each model to find out their best performance can be one direction of future work.

---

[11]https://github.com/allanj/pytorch_neural_crf
[12]https://nlp.stanford.edu/projects/glove

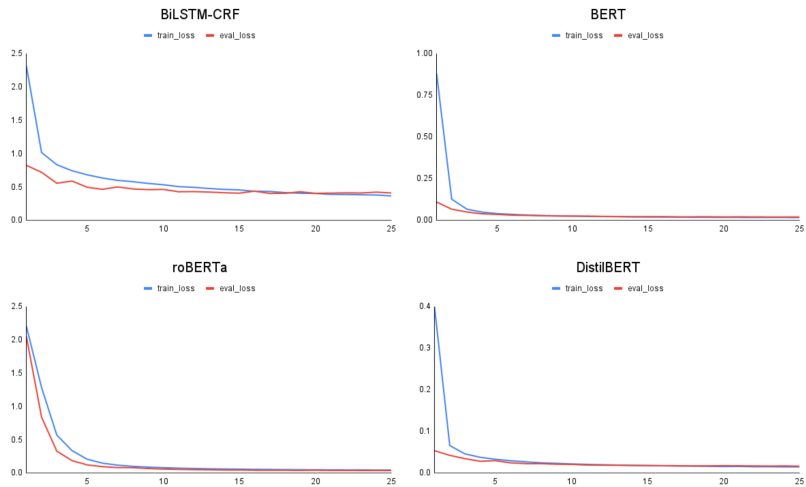

Figure 4: The training and validation loss of baseline models

## 5 CONCLUSION AND FUTURE WORK

In this work, we introduce a benchmark consisting of benchmark datasets and evaluation metrics to assess the performance of the GPT-3 model in geocoding address parsing and compare with three transformer-based models and one LSTM-based model. We create a benchmark dataset capturing 21 input errors/variations observed in real user input logs, and this dataset also contains the unique address formatting across the U.S. (i.e., 50 states and D.C). This helps to address the demand for a 'gold standard' evaluation dataset in geocoding and further guarantees that evaluation results closely reflect their performance in real-world scenarios. Our findings reveal that the Bidirectional LSTM-CRF model slightly outperforms the transformer-based models. Though the GPT-3 model's performance lags behind the other evaluated models, it shows encouraging results in address parsing using few-shot examples, suggesting room for improvement with additional fine-tuning. We aim this work to serve as a solid baseline for future development and experimental comparisons in similar geographic information retrieval-related tasks.

Future work includes (1) enhancing the evaluation benchmark dataset by capturing more input errors/variations, (2) fine-tuning the models to improve their performance to attempt to achieve SOTA performance in the given input dataset and comparing to traditional models (e.g., CRF and HMM models), and (3) extending this benchmark to be applicable to evaluate address parsing or geocoding systems in other countries, given the heterogeneity of language and address systems in different countries.

Table 4: Evaluation results of baseline models

| Address component | BiLSTM-CRF | BERT | roBERTa | DistilBERT | GPT-3 |
| --- | --- | --- | --- | --- | --- |
| House number | 0.99977 | 0.99963 | 0.99976 | 0.99970 | 0.98810 |
| Predirectional | 0.99719 | 0.99378 | 0.98996 | 0.99579 | 0.70077 |
| Street base name | 0.99241 | 0.98963 | 0.98022 | 0.99061 | 0.83853 |
| Road type | 0.99705 | 0.99345 | 0.98543 | 0.99427 | 0.88328 |
| Postdirectional | 0.96739 | 0.96680 | 0.94003 | 0.96771 | 0.42917 |
| City | 0.99399 | 0.99293 | 0.98539 | 0.99341 | 0.90404 |
| State | 0.99993 | 0.99986 | 0.99894 | 0.99991 | 0.97505 |
| Postal code | 1.00000 | 1.00000 | 0.99987 | 1.00000 | 0.97851 |
| Overall F1 | 0.99677 | 0.99545 | 0.99084 | 0.99590 | 0.90875 |
| Parsing Score | 148.37200 | 148.16378 | 147.39396 | 148.24340 | 133.32740 |